# A Structured, Probabilistic Representation of Action

**Ron Davidson**
Laboratory for Intelligent Systems
Stanford University
Stanford, CA
davidson@lis.stanford.edu

**Michael R. Fehling**
Laboratory for Intelligent Systems
Stanford University
Stanford, CA
fehling@lis.stanford.edu

## Abstract

When agents devise plans for execution in the real world, they face two important forms of uncertainty: they can never have complete knowledge about the state of the world, and they do not have complete control, as the effects of their actions are uncertain. While most classical planning methods avoid explicit uncertainty reasoning, we believe that uncertainty should be explicitly represented and reasoned about. We develop a probabilistic representation for states and actions, based on belief networks. We define conditional belief nets (CBNs) to capture the probabilistic dependency of the effects of an action upon the state of the world. We also use a CBN to represent the intrinsic relationships among entities in the environment, which persist from state to state. We present a simple projection algorithm to construct the belief network of the state succeeding an action, using the environment CBN model to infer indirect effects. We discuss how the qualitative aspects of belief networks and CBNs make them appropriate for the various stages of the problem solving process, from model construction to the design of planning algorithms.

## 1 INTRODUCTION

Real-world planning poses special challenges which early planning systems did not fully confront. Typically, the domain models upon which real-world planners rely reflect incomplete and inaccurate understanding of the domain's *ontology*, the objects and events used to describe domain states, and its *dynamics*, the underlying principles that define object-event relationships in the domain. Due to such inherent limitations of scope and accuracy, real-world planners must cope with substantive *uncertainty* as they reason about actions and formulate plans. These planners must deal adequately with uncertainty about past, present, and future states of the world including uncertainty about the occurrence of particular events in the domain. These challenges are exacerbated by the reliance on imperfect sensor information and the uncertainties introduced by the potential actions of other agents.

Early planning systems, e.g., STRIPS (Fikes and Nilsson, 1971) and NOAH (Sacerdoti, 1975) presumed availability of a complete, accurate domain representation and, consequently, produced plans that would readily fail as a result of inadequacies in the domain model. As planning researchers began to address the challenges of uncertainty several approaches emerged, including *replanning* (Wilkins, 1988), *interleaved planning and execution* (McDermott, 1978; Georgeff and Lansky, 1987), *reactive planning* (Brooks, 1986; Kaelbling, 1987; Schoppers, 1987), and *conditional planning* (Warren, 1976; Peot and Smith, 1992). These approaches reduce the negative impact of uncertainty on the quality of plans, but none represents and reasons about uncertainty explicitly.

So far, several attempts have been made to integrate uncertainty representation and reasoning techniques into planning. Markov chains are used by Christiansen and Goldberg (1990) and by Dean et al. (1993) to depict a sequence of possible actions. Kanazawa and Dean (1989) use influence diagrams similarly to Markov chains, without exploiting their structure. Several efforts focus on the design of specialized projection algorithms (e.g., Dean and Kanazawa, 1989; Hanks, 1990; Drummond and Bresina, 1990). Recently, some probabilistic and decision-theoretic systems have been implemented (Kushmerick et al, 1994; Goldman and Boddy, 1994; Haddawy and Suwandi, 1994). Most of these approaches represent uncertainty probabilistically. However, Wellman (1990a) uses qualitative probabilistic networks, while Chrisman (1992) rejects Bayesian probability in favor of Belief Functions.

Some of these probabilistic approaches contrast strongly with the more classical AI approaches. While it is agreed that transition matrices provide a complete representation for actions' effects and readily support probabilistic temporal projection, this representation poses extreme challenges. On the practical side, a daunting amount of assessment may be needed to construct a complete Markov transition matrix, and inference with matrices cannot easily support queries about specific properties of states. Perhaps more importantly, it remains unclear how, if at all, matrices can be incorporated into anything like conventional planning techniques.

In contrast, our approach to planning under uncertainty aims to incorporate a suitable treatment of uncertainty within a more conventional overall planning process. As a first step in this research we have focused on the representation of actions as operators that probabilistically transform states by specifying probabilistic relationships among their descriptive elements. We build our framework on belief networks (Pearl, 1988) and a variant thereof to represent conditional probability distributions. Contrary to the probabilistic planners mentioned above, we do not make the STRIPS assumption that anything that is not explicitly part of the action's model remains unchanged. In this respect, our work may be compared to recent research on action representation in belief networks (Goldszmidt and Darwiche, 1994; Pearl, 1994). We propose a simple projection algorithm that uses graphic operations to construct the probabilistic model that represents the direct and indirect effects of an action. Our action representation combines qualitative and quantitative description in order to support reasoning under uncertainty in the context of more conventional qualitative methods for reasoning about action.

## 2 MODELING WORLD STATES

The following example will motivate our presentation of this research: a robot is secretly attempting to fetch an object from a room in the WhiteWaterGate office building. The robot must avoid detection. It has a partial description of the object, including its location, size and weight. Upon locating the object, the robot is now reasoning about picking it up, but since it must accomplish the task without being detected, it also must assess the possibility of activating an alarm. Thus, the appropriateness of the pickup action depends upon the robot's model of the state of the world when the action is attempted, and its model of how the action, if undertaken, might affect critical distinctions[1] in that state.

Since the robot is uncertain about most aspects of the environment and the exact effects of actions, a probabilistic representation is called for. One possible approach would be to model the robot's situation as a probability distribution over all the possible states, with a conditional probability distribution for the states that may result from executing the action. An action is, thus, represented by a state-transition probability matrix. However, this representation technique hides important qualitative information about relationships in the domain that may be important to planning. A representation that provides explicit information about independence and conditional independence among distinctions in the model can make descriptions more compact and expressive while helping to improve efficiency of inference algorithms. We, therefore, represent world states (or simply states) by belief networks (Pearl, 1988) which depict the factorization of a joint probability distribution in a graphical manner:

**Definition 1:** Let $P(D)$ be a joint probability distribution over a set of distinctions $D$. Let $G$ be a directed acyclic graph $(D, R)$, where $D$ is a set of nodes that correspond to the distinctions[2], and $R$ is a set of directed arcs. Let $(d_1,...,d_n)$ be a node ordering consistent with $G$, i.e., if $(d_i, d_j) \in R$ then $i < j$. Let $\pi(d)$ be the set of parents of node $d$ in $G$: $\pi(d) = \{c \in D \mid (c, d) \in R\}$. Then, we say that $G$ is a *belief network* for $P$ if for every node $d_i \in D$, $P(d_i \mid d_1,...,d_{i-1}) = P(d_i \mid \pi(d_i))$, and no arc can be removed from $G$ without violating this factorization. We often attach the conditional probability distribution $P(d_i \mid \pi(d_i))$ to the node $d_i$ and view the network as representing $P$.

**Definition 2:** A *state of the world* $W$ is a specification of the values of all the distinctions of interest at a snapshot of the real world. A joint probability distribution $P_W(D)$ is a *state model* if it reflects the uncertain beliefs of an agent about $W$. We often use the corresponding belief network as a model for $W$.

The belief network in figure 1 shows that the robot has some prior information about possible size and location of the object, and believes that object weight is related to size. (I.e., information about one of these features provides information about the other.) However, both these features are independent of (provide no information about) the object's location. The alarm has three independent sensor sources—light, sound, and motion. The robot believes that the value of each of these is probabilistically related to the alarm activation. Also, the chance of being discovered by a guard is believed to be related to the alarm activation.

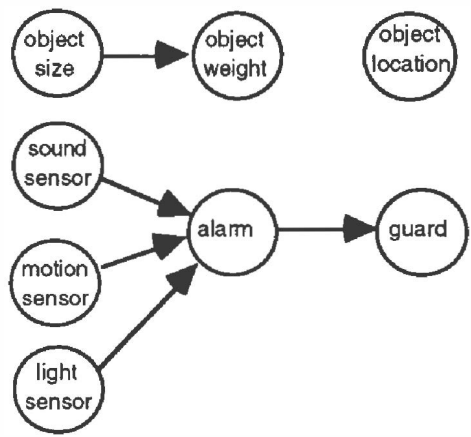

Figure 1: A World State

Several aspects of belief networks deserve mention. First, the most important qualitative information conveyed by belief networks lies in the arcs that are missing, i.e., in the

---

[1] A *distinction* is a predicate or a random variable describing some property of the task domain.

[2] For this discussion, we need not distinguish between a node in the graph and the model distinction it represents.

*independence assertions.* In figure 1, one can readily see the independence of object location from all other distinctions and the independence among the alarm sensors. This can be seen without examining quantitative information about the joint distribution through the graphical criterion of *d-separation* (Geiger et al, 1990). Second, arcs do *not* necessarily imply causality. Some arcs may be reversed without affecting the rest of the network (e.g., between object size and weight). In fact, *any* arc can be reversed, with possible modifications to the rest of the network (Shachter, 1986). Third, a node may be a deterministic function of its predecessors (e.g., the alarm is activated if, and only if at least one sensor is on), but generally, a node remains probabilistic even if predecessor values are known (e.g., the robot may not be discovered even if the alarm is activated).

## 3  REPRESENTING ACTIONS

World state models describe relationships within a specific state. We now model actions taking the standard view of actions as transitions between consecutive states. Thus, given the state of the world preceding the action, an action model defines probabilistic constraints on the state succeeding the action. In other words, an action could be described by a conditional probability distribution for the state succeeding the action given the state preceding the action. We do not wish, however, to require that every action be specified by a full probability matrix, for reasons stated above. We seek a compact representation that states only the intrinsic properties of an action.

Before presenting the formal definition, consider the action of our robot picking up the object. The robot is uncertain whether the object will end up in its grasp. Thus, one probabilistic effect of the action is object location—the object may remain in the same location (if it is too heavy to lift, say), the object may fall to the floor (if its size makes it awkward to carry to the loading bay), or, hopefully, the object's new location may indeed be the robot's bay. Another uncertain effect is the activation of the alarm by triggering one of the sensors. Thus, some distinctions in the model qualify the effects of the action, while other distinctions are affected by the action. So, for any model $M$ of action $A$, we denote the set of qualifying distinctions in the state preceding $A$ by $qual_M(A)$, and the set of the directly affected distinctions in the succeeding state by $eff_M(A)$. The notion of direct effects is captured by the definition below. In the sequel we omit the model designation $M$ unless we discuss different models for the same action.

The following definition for $qual(A)$ and $eff(A)$ express our intent that these sets are the minimal ones required for an adequate representation of $A$. The requirements we impose on these sets are in line with Wellman's characterization of probabilistic actions (1990).

**Definition 3:** Let $P$ and $S$ be the sets of distinctions that correspond to the states that precede and succeed an action (respectively). Then, for every action $A$, $qual(A) \subseteq P$ and $eff(A) \subseteq S$ are defined to be the minimal sets of distinctions such that:
- $eff(A)$ is independent of $P-qual(A)$ given $qual(A)$ and the fact that $A$ was performed
- $S-eff(A)$ and $P-qual(A)$ are independent of $A$ given $qual(A)$ and $eff(A)$.

The last requirement says that given $qual(A)$ and $eff(A)$, knowledge that $A$ was performed does not convey any additional information about any distinction in the model[3]. Note that even if we have the full transition matrix (or the conditional probability distribution) that corresponds to action $A$, we cannot derive $qual(A)$ and $eff(A)$ directly, as our ability to distinguish the execution of action $A$ depends on the other possible actions that could have been performed. Thus, $qual(A)$ and $eff(A)$ need to be specified by the domain expert. Once they are specified, $P(eff(A) | qual(A))$ can be considered as the compact model of the action.

**Definition 4:** Let $P(E | Q)$ be a conditional probability distribution. Then $P$ is a *model for action $A$* if $Q=qual(A)$ and $E=eff(A)$.

As actions are defined in terms of conditional probability distributions, and in light of our emphasis on qualitative representation, we now define a variant of a belief network—a *conditional belief net* (CBN)—to graphically display a conditional distribution.

**Definition 5:** Let $P(E | Q)$ be a conditional probability distribution of $E$ given $Q$ (where $Q$ and $E$ are disjoint sets of distinctions). Let $B$ be a directed acyclic graph $(Q, E, R)$, where $Q \cup E$ is a set of nodes and $R \subseteq (Q \cup E) \times E$ is a set of directed arcs. Let $(Q, e_1,...,e_n)$ be a node ordering consistent with $B$ and let $\pi(e)$ be the set of parents of node $e$ in $B$. We say that $B$ is a *conditional belief network* (CBN) for $P(E | Q)$ if for every node $e_i \in E$, $P(e_i | Q, e_1,...,e_{i-1}) = P(e_i | \pi(e_i))$, and no arc can be removed from $B$ without violating this factorization.

Note that the order of the nodes in $Q$ is irrelevant as there are no arcs going into nodes in $Q$. For these nodes, only their list of possible values needs to be specified. Since a CBN $(Q, E, R)$ does not specify a probability distribution for the nodes in $Q$, we often term $Q$ as the set of *free* distinctions of the CBN, while the set $E$ is the set of *bound* distinctions, as they are constrained by a probability distribution.

A model of the pickup action is given in figure 2. The nodes surrounded by a rounded rectangle represent the affected distinctions in the succeeding state. A special node marks the action's name, with an outgoing arc into the effects. Although this extra node is redundant, it will prove useful in representing the combined effects of simultaneous actions and for the construction of decision models for sequences of actions.

---

[3] Moreover, $qual(A)$ and $eff(A)$ render $A$ independent of any distinction in the past or in the future.

How should we interpret figure 2? Object location is affected by the action. Its value in the succeeding state depends (probabilistically) on its location before the action and on the object's size and weight. The model relates the status of the sound sensor after the action to its status before the action (if it was on, it will probably remain on), and to the object location after the action (since if it falls to the floor, it makes noise). Note that the alarm activation is not specified as a possible effect of the action. It was not judged to be a direct effect of the action.

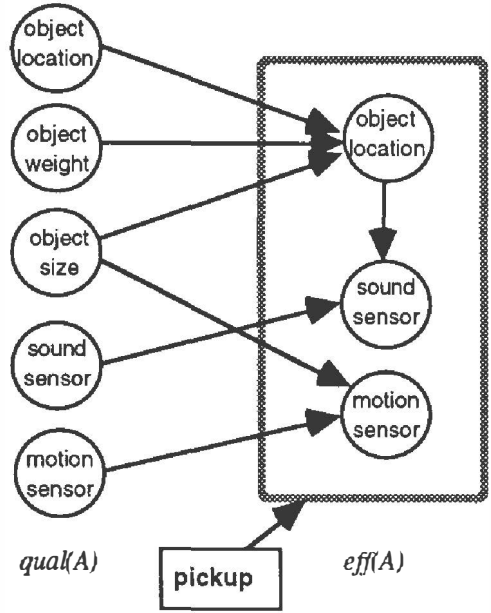

Figure 2: An Action Model

## 4 THE ENVIRONMENT MODEL

A belief network represents the relationships between distinctions within a particular state. Effective reasoning about action requires that we understand and represent contingent relationships that extend beyond any specific state. These relationships are inherent in the environment or the system we model, and we can use them to achieve goals (e.g., to induce rain we can seed clouds) and to avoid undesirable side effects (e.g., if we make noise, we might trigger the sound sensor).

The knowledge that characterizes invariant relationships in an uncertain system or environment is best described by conditional probability distributions (in much the same way that dynamic systems are described by differential equations). These conditional relations are contingent upon inputs whose distribution may not be known at modeling time. For example, we do not know the status of the sound sensor in general, but we may know the conditional probability of alarm status given sensors' status, and this conditional relation holds for every possible state of the world.

We collect all the relationships that are expected to hold in every state into a conditional probability distribution, designated as *the Environment Model*. While these relations are not changed by any action the agent may take, they may change in particular states due to *observations*. Thus, if the agent hears the alarm in a certain state, it is no longer probabilistically related to the status of its sensors in that state.

**Definition 6:** Let $D$ be the set of distinctions in terms of which states are described. Let $(F;H)$ be a partition of $D$: $D=F\cup H$, $F\cap H=\emptyset$ such that the set of probabilistic relationships $P(H|F)$ is believed to hold in every state. The conditional distribution $P(H|F)$ (and its corresponding CBN) are called *the Environment Model*.

The set of the free distinctions $F$ is not constrained by the Environment Model. The bound distinctions in the set $E$ are constrained to the same distribution in every state (unless observations are available).

Figure 3 displays the Environment Model for the robot's example. The free distinctions, for which no probabilistic information is specified, are the object location and the three sensors. The distribution of all other distinctions (the bound distinctions in the rounded rectangle) is believed to be the same in all states.

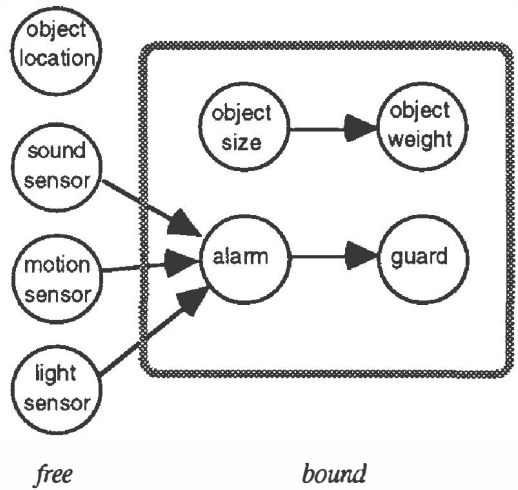

*free*      *bound*

Figure 3: The Environment Model

The Environment Model can be transformed into a world state model by adding a probability distribution for the nodes in $F$. This transforms the CBN into a belief network, from which we can derive the marginal probability of every distinction of interest. We can draw an analogy between the Environment Model and a set of differential equations that defines a physical system. The free distinctions correspond to the boundary conditions of the system, whose specification permits one to calculate the entire behavior of the system.

## 5 STATE PROJECTION

Temporal projection is about inferring the state of the world after an action is taken. We now present a projection algorithm exploiting our CBN-based representations.

The state that succeeds an action depends, in general, on the preceding state and on the action's model. The algorithm below constructs a belief network that combines both the preceding and the succeeding states, and the relations between them. The model of the succeeding state can then be extracted from the combined network.

**Algorithm Project_State**

Input: A belief network $G_p$ over $D$ representing the preceding state and a CBN $B$ over $Q \cup E$ that represents an action $A$

Output: A belief network $G_{ps}$ over $D \cup E$ representing the relations between the preceding and the succeeding states, and a belief network $G_s$ over $D$ representing the succeeding state

1. Initialize $G_{ps}$ to the network $G_p$ of the preceding state.
2. Augment $G_{ps}$ by the CBN $B$ in the following way:
   A. Coincide the free nodes $Q$ (in $B$) with the nodes with the same label in $G_{ps}$.
   B. Add the bound nodes $E$ to $G_{ps}$ as succeeding-state nodes. These nodes inherit their probabilistic information (their incoming arcs) from $B$.
   C. Let $F$ be the distinctions in $G_p$ that correspond to $E$ (i.e., the distinctions in the preceding state that are affected by the action.). Let $K$ be the set of descendents of $F$ in $G_p$: $K = \bigcup_{f \in F} des(f) - F$, where $des(f)$ are nodes to which there exists a directed path in $G_p$ from $f$. The nodes in $K$ are duplicated, so that a new copy of these nodes is added to $G_{ps}$ in the succeeding state. The new nodes carry their probabilistic information (their incoming arcs) from the preceding state in $G_{ps}$. Whenever possible, these arcs originate from a new copy of a node, i.e., from the succeeding state.
3. $G_s$ can constructed from $G_{ps}$ by computationally removing the set $F \cup K$ of nodes from the preceding state of $G_{ps}$. The resulting belief network $G_s$ represents the succeeding state.

A node is computationally removed (step 3 in the algorithm) following the procedure described in (Shachter, 1986). The procedure essentially averages out a node, so that the distribution of its children is properly updated. For most purposes, it is advised not to perform step 3, and use the combined network $G_{ps}$ for inference, so as not to lose relevance information across states. In other words, if the purpose of projection is to construct a probabilistic model that corresponds to a sequence of actions, step 3 should not be performed. (In this case, step 2.C requires that new arcs originate from the *most recent* copy of a node). If, however, the purpose of projection is to calculate the marginal probabilities for the distinctions in the succeeding state, step 3 is necessary.

Figure 4 depicts the combined network for the pickup action of our robot example. The input to the algorithm is the preceding state as depicted by figure 1, and the action model in figure 2. The nodes in $E$ (the direct effects: object location, sound sensor and motion sensor) are added to the network of the preceding state, with their incoming arcs. These nodes are distinguished by the rounded rectangle, while all the succeeding-state nodes are annotated by s under their label. In the next step, the nodes in $K$ (the *indirect* effects: alarm, guard) are added to the succeeding state, with their incoming arcs originating from succeeding-state nodes, if possible. For example, the alarm node has two incoming arcs from new nodes, but the arc from the light sensor node comes from the preceding state.

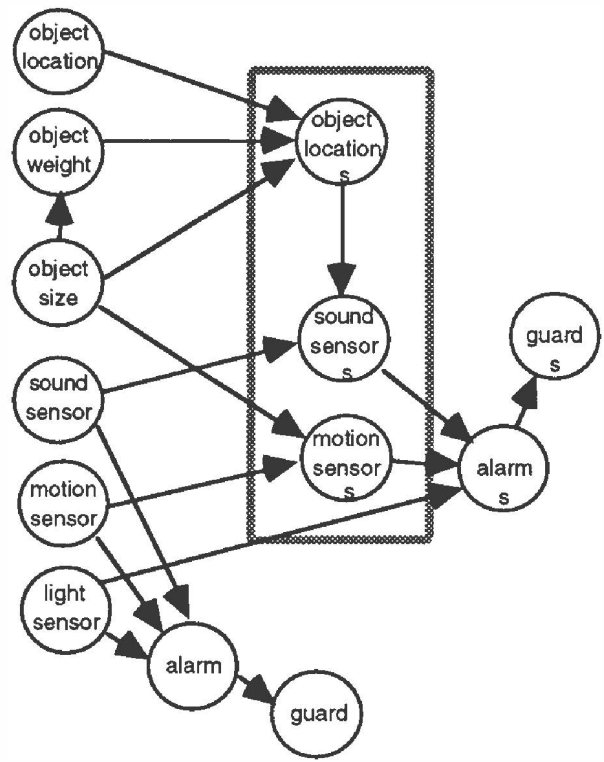

Figure 4: State Projection

Note that the nodes in $D - F - K$ are not affected by the action and are not copied into the succeeding state. We could, of course, copy them as deterministic identity functions of their value in the preceding node.

We can project the succeeding state from the combined network by computationally removing all the nodes $F \cup K$ in the preceding state. The node removal algorithm (Shachter, 1986) assures that the probability distribution of the remaining nodes is consistent with the agent's beliefs. Figure 5 depicts the succeeding state network $G_s$. We can now ignore the annotation by s since the network includes exactly all the distinctions in $D$ of the succeeding state.

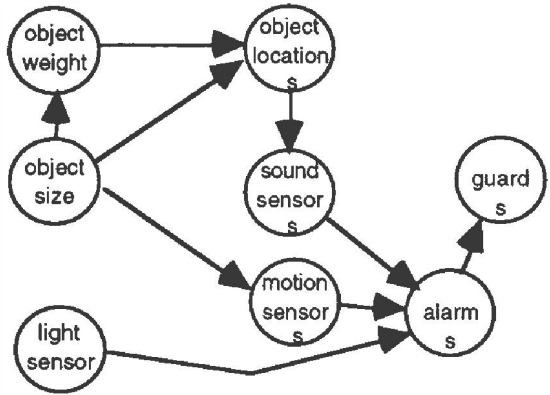

Figure 5: The Projected State

## 6 PROJECTING STATES CORRECTLY

Note that the network in figure 5 (the succeeding state) has a different structure from the one in figure 1 (the preceding state). In particular, object size became probabilistically relevant to object location and motion sensor, and similarly, information about the sound sensor is relevant to object location. Thus, the action introduced informational relationships between entities in the domain.

There is a potential problem with our very flexible approach. Suppose next, our robot performs an action that directly affects the size of the object (e.g., cutting it into half). Our projection algorithm would infer an indirect effect on both the object location and the motion sensor. Similarly, the robot may now have a silent way to move the object, so a subsequent change in object location need not necessarily have the status of the sound sensor as an indirect effect. The main point here is that these informational relations that are introduced by the action do not necessarily persist from one state to another. If a subsequent action affects a distinction $d$ that is associated with another distinction $e$, the new value $d_s$ in the succeeding state need not be related to $e$ or $e_s$.

In general, the algorithm as outlined above is sensitive to the direction of the arcs in the network of the preceding state. Since every arc can be reversed employing the graphical equivalent of Bayes' rule (Shachter, 1986), our projection algorithm may yield different results given different graphical representations of the preceding state.

These difficulties should not come as a surprise. As Pearl (1994) notes: "While (a probability distribution) tells us everything about responding to new observations, it tells us close to nothing about responding to external actions" (p. 204). Thus, the model of the preceding state cannot be sufficient. Additional assumptions must be made.

To resolve the problem we turn back to the Environment Model. Recall that the Environment Model represents the probabilistic relations that are believed to hold in every world state. These are the only relations that we must guarantee to hold in the succeeding state. In other words, these are the relations that *persist* from state to state, unlike the other relations in a state, which are purely ad-hoc.

We modify slightly the projection algorithm from the previous section to rely on the Environment Model to derive the indirect effects of an action (instead of on the network of the preceding state). The modified algorithm accepts as input the Environment Model $V$ too, and it differs only in the method of identifying and deducing the indirect effects.

### Algorithm Project_State (modified)

Input: add the Environment Model $V$ to the original inputs
Perform steps 1-2.B as in the original algorithm.

2.C. Let $F$ be the distinctions in $V$ that correspond to $E$ (i.e., the distinctions in the Environment Model that are directly affected by the action). Let $K$ be the set of descendents of $F$ in $V$: $K = \bigcup_{f \in F} des(f) - F$, where $des(f)$ are nodes to which there exists a directed path in $V$ from $f$. The nodes in $K$ are duplicated, so that a new copy of these nodes is added to $G_{ps}$ in the succeeding state. The new nodes carry their probabilistic information (their incoming arcs) from the $V$. These arcs originate from the most recent copy of a node.

As before, nodes can be removed to derive a belief network that represent the succeeding state solely.

For the robot's example above, the combined network for the pickup action would be the same with the modified algorithm (with the Environment Model in figure 3 as an additional input) as it was with the original algorithm. That is, the modified algorithm projects the same succeeding state as in figure 5. However, if a subsequent action changes the location of the object independent of the sensors, the modified algorithm would not infer the sound sensor (and the rest of the alarm system) as indirect effects. They would retain their values from the preceding state.

Without imposing any restriction on action models, a potential conflict might arise between the direct effects of actions and relations in the Environment Model. If a bound distinction $e$ in $V$ is a direct effect of an action $A$ (i.e., if $e \in \textit{eff}(A)$), then it is not clear whether $e$ should be defined in the succeeding state as in the action model or as in the Environment Model (as we assumed that $e$ is bound to the distribution in $V$ in *all* states). We gave absolute priority to the action model in the projection algorithm, but this is an arbitrary decision.

As we discuss in section 7, we believe that such a conflict is a result of poor modeling, and we therefore introduce the following restriction on action models:

**Definition 7:** A model $M$ for action $A$ is said to be *compatible with the Environment Model $V$* if the direct effects $\textit{eff}_M(A)$ are restricted to the free distinctions of $V$.

From now on, we assume that all action models are compatible with the Environment Model. It is not difficult to verify to that with this restriction on actions, the modified projection algorithm preserves consistency of state models with the Environment Model:

**Definition 8:** A state model W is *consistent with the Environment Model* V if for every bound distinction $h$ in V, $P_V(h|\pi_V(h)) = P_W(h|\pi_V(h))$, where $\pi_V(h)$ is the set of parents of $h$ in V.

**Proposition 1:** If the model of the preceding state is consistent with the Environment Model V, and the action model is compatible with V, then the model of the succeeding state, as derived by the modified projection algorithm, is consistent with V too.

Note that the original projection algorithm satisfies this proposition as well. This is to say that proper projection cannot be judged only by the consistency of the resulting state with the relations in the Environment Model. Rather, it is the appropriate handling of the distribution of the free distinctions that makes the difference.

Note that we are unable to assert an objective criterion for the *correctness* of the projection algorithm. We can only verify that the projection procedure is consistent with the assumptions we make (proposition 1), argue for the reasonability of these assumptions and test them.

Our primary assumption is that there are probabilistic relations between entities in the real world that are expected to hold in every state (unless modified by an observation of one of the relevant entities). Moreover, we assume that all other probabilistic relations are just informational relationships that hold only for specific entities at the specific time when they are asserted.

The persistent relationships in the Environment Model are reminiscent, of course, of causal networks (Pearl, 1988). These are belief networks in which all arcs are assumed to have a causal meaning. This strong assumption is made by (Pearl, 1994) and (Goldszmidt and Darwiche, 1994) to permit reasoning about action and change within the framework of belief networks. Indeed, as Druzdzel and Simon (1993) note: "The effect of a structural change in a system cannot be induced from a model that does not contain causal information" (p. 4). We delay to a forthcoming paper the discussion of the causal interpretation of our Environment Model, as well as a comparison to the contemporary works mentioned above and that of Heckerman and Shachter (1994).

## 7  PROPERTIES OF ACTION MODELS

We now briefly discuss the restriction we imposed on actions to affect only free distinctions in the Environment Model, and the uniqueness properties that result.

Without the compatibility restriction, an action A could have models with different effects sets $\textit{eff}_M(A)$ that would be projection-wise equivalent: they would result in the same succeeding state. This would happen if some effects set included non-direct effects as part of the action model, i.e., if a modeler fails to distinguish the genuine direct effects from those that result from relationships in the environment. Compatible models can never be equivalent for *every* preceding state, as any free distinction not in $\textit{eff}_M(A)$ persists when M is used for projection.

The compatibility restriction is useful beyond avoiding conflicts between action models and the Environment Model, and promoting uniqueness of effects. It also facilitates modeling (knowledge acquisition) in the following way: if a need arises to model a bound distinction $h$ as an effect, the Environment Model V is not complete, since it ignores a possible way to affect $h$. For example, if V specifies $P_V$(grass|rain) and we identify a way to affect grass (e.g., by turning the sprinkler on), we conclude the V is incomplete as it neglects the influence of sprinkler on grass. Instead, V should specify $P_V$(grass|rain,sprinkler).

In general, the sets $\textit{qual}_M(A)$ are not unique. For example, if V contains a deterministic bound distinction $h$, then any model M with $h \in \textit{qual}_M(A)$ has an equivalent model M' with $\pi_V(h)-\{h\} \in \textit{qual}_{M'}(A)$. If, however, all distributions are strictly positive, the intersection property (Pearl, 1988, p. 84) can be used to define a minimal unique qualifiers set $\textit{qual}(A)$.

## 8  SUMMARY AND DISCUSSION

We have presented a scheme for modeling the states of the world and the effects of actions in a probabilistic fashion. We use belief networks to represent world states, and *conditional belief nets* to represent actions and environmental contingencies. Our representation scheme enjoys the advantages of qualitative modeling and the precision of quantitative models. The representation supports efficient projection and prepares for intelligent planning by emphasizing the properties and the structure of states. The framework is flexible, and can be extended in several ways (Davidson, 1994).

Our work is influenced by the classical action-representation schemes such as STRIPS (Fikes and Nilsson, 1971) and the situation calculus (McCarthy and Hayes, 1969). Like STRIPS, our action models depict the relationships among preconditions and effects, though we relate them probabilistically. STRIPS suffers from the need to explicitly specify the truth value of all formulas that could possibly be affected by an operator. Attempts to eliminate the problem (Wilkins, 1988) allow a set of basic formulas to appear in add/delete lists, from which all other formulas are calculated. Our approach is similar: action models may affect only the free distinctions in the environment, while all other distinctions are conditioned upon the free ones, and cannot be affected directly.

Any scheme similar to the situation calculus suffers from the frame problem. Our solution to the extended prediction problem is similar to STRIPS': we assume persistence of whatever is not affected, but we handle indirect effects. Moreover, our framework addresses the qualification problem by acknowledging that a model can never exhaust the qualifying distinctions in the real world, and therefore, all effects are probabilistic.

Our work on this scheme continues. We believe that the framework will prove appropriate for decision-theoretic planning, whereby maximum expected value is the criterion for the optimality of plans. We are now at work to introduce levels of abstraction into this representation scheme, and to devise a hierarchical planning algorithm for this approach.

### Acknowledgments

We wish to thank Ross Shachter and Eric Johnson for useful discussions, Bill Poland for reviewing an earlier draft, and the anonymous referees for their comments. The remaining errors are solely ours, of course. Work reported here was supported by funds from grant no. N00014-93-1-0324 from the Office of Naval Research.